\documentclass[letterpaper, 10 pt, journal,twoside]{IEEEtran}
\usepackage{amsmath,amssymb,amsfonts}
\usepackage{balance}
\usepackage{bm}
\usepackage{hyperref}
\usepackage{color}
\usepackage[table,xcdraw]{xcolor}
\usepackage{tabularx}
\usepackage[caption=false,font=normalsize,labelfont=sf,textfont=sf]{subfig}
\usepackage{booktabs}
\usepackage{graphicx}
\usepackage{multirow}
\usepackage{tikz}
\usepackage{pgfplots}
\usepackage{siunitx}
\usetikzlibrary{bending, external}
\usepackage{threeparttable}

\graphicspath{{./Figures/}}
\DeclareGraphicsExtensions{.pdf,.png,.jpg,.eps,.svg}

\IEEEoverridecommandlockouts
\pgfplotsset{compat=1.16} 

\title{Threat-Aware UAV Dodging of Human-Thrown Projectiles with an RGB-D Camera
}

\author{Yuying Zhang, Na Fan, Haowen Zheng, Junning Liang, Zongliang Pan, Qifeng Chen, and Ximin Lyu
\thanks{
\textit{(Yuying Zhang and Na Fan are co-first authors.) (Corresponding author: Ximin Lyu.)}
}

}

\begin{document}
\maketitle

\begin{abstract}

Uncrewed aerial vehicles (UAVs) performing tasks such as transportation and aerial photography are vulnerable to intentional projectile attacks from humans. Dodging such a sudden and fast projectile poses a significant challenge for UAVs, requiring ultra-low latency responses and agile maneuvers. Drawing inspiration from baseball, in which pitchers' body movements are analyzed to predict the ball's trajectory, we propose a novel real-time dodging system that leverages an RGB-D camera. Our approach integrates human pose estimation with depth information to predict the attacker's motion trajectory and the subsequent projectile trajectory. Additionally, we introduce an uncertainty-aware dodging strategy to enable the UAV to dodge incoming projectiles efficiently. Our perception system achieves high prediction accuracy and outperforms the baseline in effective distance and latency. The dodging strategy addresses temporal and spatial uncertainties to ensure UAV safety. Extensive real-world experiments demonstrate the framework's reliable dodging capabilities against sudden attacks and its outstanding robustness across diverse scenarios. Project page: \url{https://threat-aware-dodging.github.io/}

\end{abstract}
\vspace{-0.2cm}
\begin{IEEEkeywords}
\vspace{-0.5pt}
Aerial systems: Perception and Autonomy, UAV safety, Human pose estimation, Projectile dodging
\end{IEEEkeywords}

\vspace{-9pt}

\section{INTRODUCTION}	

\IEEEPARstart{T}{he} rapid advancement of uncrewed aerial vehicles (UAVs) and their supporting infrastructure has significantly expanded the UAV market, enabling diverse applications such as aerial imaging, last-mile delivery, and air traffic management ~\cite{2020SR_Dynamic_Falanga, wang2025safe}. To meet the demands of these complex tasks, modern UAVs are increasingly equipped with autonomous modules for environmental perception, navigation, and obstacle avoidance. Despite these advances, UAVs often fail to cope with sudden human-initiated attacks. Recent reports have documented cases where crowds at public events throw projectiles to disrupt UAV operations~\cite{UAVAttackReport_20230928_Reddit, UAVAttackReport_20250204_AZFamily}, posing significant threats to their safety and public security. Consequently, there is an urgent need for robust strategies to counter human-initiated attacks involving fast-moving projectiles.

Developing robust UAV systems capable of rapid responses to sudden human-initiated attacks remains a critical and unresolved research problem. Dodging such projectile threats involves overcoming several challenges:
(1) \textit{Perception Latency:}
Projectiles often emerge suddenly at close range, leaving a narrow time window for detection and dodging. Therefore, minimizing the delay between sensing and control is crucial while maintaining high prediction accuracy to ensure effective avoidance.
(2) \textit{Inaccurately Predictable Threats:}
The unpredictable timing of human-initiated attacks, combined with the complex dynamics of thrown projectiles influenced by environmental factors such as wind, presents significant challenges in accurately modeling projectile trajectories.
(3) \textit{Computational Constraints and Cost-effectiveness:}
UAV hardware must support real-time computing capabilities while remaining cost-effective to facilitate widespread deployment. Furthermore, the integration of affordable, high-performance sensors and processors is crucial for enabling scalable and reliable threat-aware projectile dodging.

\begin{figure}[t]
\vspace*{-0.2cm}
	\centering
	\includegraphics[width=0.43\textwidth]{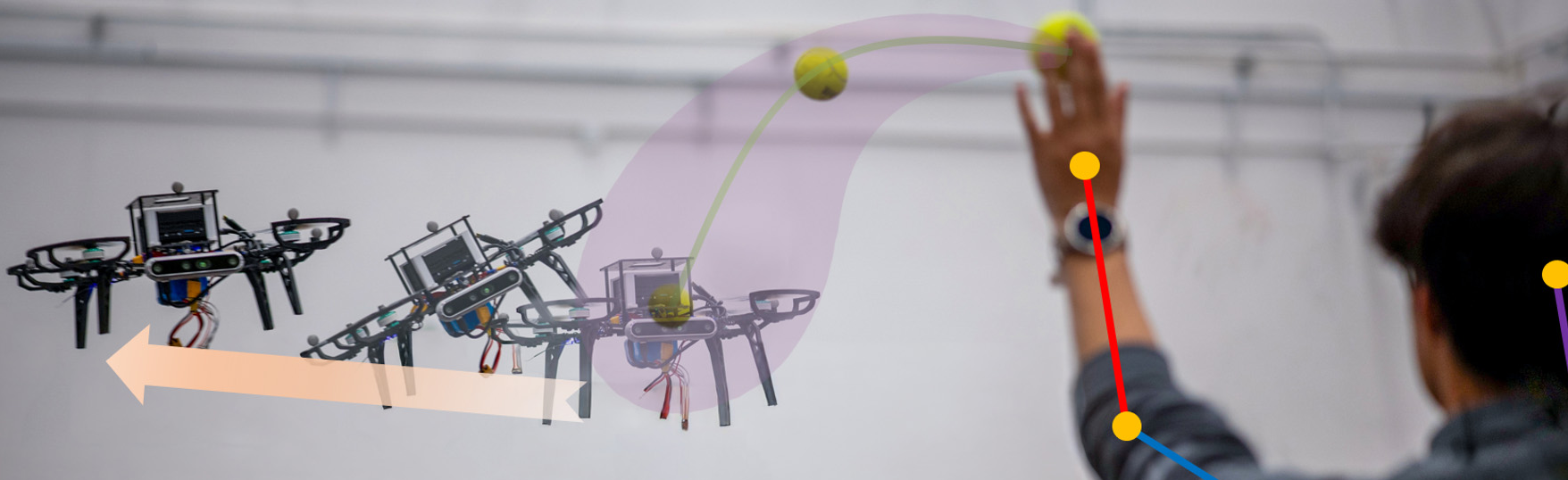}
    \vspace{-0.2cm}
	\caption{Upon detecting the human's intention to attack with a tennis ball, the UAV executes a rapid dodging maneuver to avoid the sudden, fast projectile.
	\label{fig:cover}}
\vspace*{-0.8cm}
\end{figure}

To address these challenges, we draw inspiration from baseball, where batters facing high-speed pitches often struggle to hit the ball if they react only after the pitcher's release. Instead, they anticipate the pitch by observing the pitcher's posture. Motivated by this analogy, we propose an onboard real-time predictive dodging framework based on an RGB-D camera. Unlike conventional approaches that track projectiles post-launch~\cite{2022RAL_FastDodging_Lu}, our strategy predicts trajectories by analyzing human poses, which enables early anticipation even before the projectile becomes visible or is released. This capability significantly enhances early motion planning. Moreover, since human skeletal features are substantially larger than the projectile, our method supports longer-range predictions. To mitigate uncertainties in attack timing and trajectory, we introduce an ivory-shaped uncertainty model that expands the hazard zone of the projectile, allowing the UAV to execute emergency dodging maneuvers under hazardous conditions. By leveraging a large-scale pretrained human pose estimation (HPE) model~\cite{2023CoRR_RTMPose_Jiang}, our approach achieves robust generalization to diverse attackers and projectiles without the need for retraining. Extensive real-world experiments demonstrate that our method, when deployed on hardware with limited computational resources, achieves significant improvements in detection latency, prediction accuracy, and dodging success rate (SR).

Our contributions can be summarized as follows:
\begin{enumerate}
\item \textbf{Threat-Aware Dodging System}:

We propose a novel threat-aware framework for UAVs to dodge sudden attacks from humans using only an RGB-D camera. Our approach predicts projectile trajectories by recognizing human poses and executes dodging maneuvers based on an uncertainty-aware dodging strategy to adjust the UAV trajectory.

\begin{enumerate}
\item [1A.] \textbf{Pose-Aware Projectile Trajectory Prediction (PAPT)}:
We develop a pose-aware framework for real-time perception and prediction of projectile trajectories, integrating human pose estimation with motion modeling to achieve a 6-meter effective detection range and an average detection latency of 26.4 ms on CPU-only hardware.
\item [1B.] \textbf{Uncertainty-Aware Dodging Strategy (UAD)}:
To address the challenges in accurately predicting attack timing and projectile trajectory, we propose an uncertainty-aware dodging strategy. Our approach constructs a joint uncertainty region that encompasses multiple surviving projectile trajectories and incorporates a penalty mechanism to optimize the UAV's current trajectory.
\end{enumerate}

\item \textbf{Extensively Tested in Real-World Environments}:
We conduct extensive real-world evaluations under diverse conditions, including varying speeds, launch angles, distances, throwers, lighting and occlusion
settings, projectile types, and multiple simultaneous threats, providing compelling evidence of the framework's effectiveness and robustness.

\end{enumerate}

\vspace{-0.25cm}

\section{Related Work}
\label{sec:related_work}

Thrown projectiles represent a type of sudden and fast dynamic obstacle. Protecting UAVs from human-initiated attacks involving such projectiles requires two essential capabilities: early detection of human threats and rapid dodging of the projectiles. Early dynamic obstacle avoidance methods, designed primarily for slower-moving pedestrians~\cite{2017RAL_RealTime_Nageli,2019RAL_CCNMPC_Zhu,2020ICRA_Robust_Lin,chen2022real,xu2025intent}, are inadequate for such intentional attacks due to the critical demands of low-latency perception and robust generalization. Consequently, recent research has shifted focus to dodging sudden, fast projectiles, often leveraging advanced sensing and intention prediction techniques. 

\vspace{-0.25cm}
\subsection{UAV Dodging of Sudden Fast Projectiles}
\label{subsec:controller}
There is a growing interest in UAVs dodging the sudden, fast projectile, where prediction-response latency becomes critical. 
This research trend emerges as event cameras gain popularity, with many approaches utilizing their low-latency, asynchronous detection of per-pixel brightness changes over fixed-rate capture. 
Mueggler et al.~\cite{2015ECMR_Towards_Mueggler} pioneered the use of probabilistic models and extended Kalman filters for tracking the projectile using event-based stereo vision. However, their analysis was primarily limited to drone hovering under ideal conditions. Subsequent studies, such as ~\cite{2020SR_Dynamic_Falanga,2018IROS_BetterFlow_Mitrokhin,2020ICRA_EVDodgeNet_Sanket} and~\cite{2019RAL_HowFast_Falanga}, addressed challenges like noise and motion compensation in event cameras during dynamic motion. He et al.~\cite{2021IROS_FAST_He} proposed an asynchronous fusion of event and depth camera data, which improves the accuracy of 3D trajectory prediction for moving objects. Nevertheless, event cameras are typically bulky, noisy, expensive, and sensitive only to moving objects, which limits their ability to perceive environments fully and restricts their practical applications. Consequently, researchers have explored alternative solutions. Kong et al.~\cite{kong2021avoiding} demonstrated a method that utilizes 3D LiDAR sensors combined with onboard processing to achieve high-frequency, safe navigation in cluttered and unknown environments, effectively avoiding small dynamic obstacles. Despite these strengths, the approach fundamentally treats dynamic obstacles as static entities, and LiDAR sensors are also heavy and expensive. Lu et al.~\cite{2022RAL_FastDodging_Lu} adopted an RGB-D camera with YOLO-FastestV2~\cite{YoloFastestV2} for ball detection. 
However, their approach requires training on 2000 tennis ball images and fixed attack distances due to camera resolution limitations, hindering its generalization to new scenarios. 
Recent work~\cite{chen2023trajectoryformer} uses a transformer architecture to predict object motion from multi-frame LiDAR sequences with high 3D tracking accuracy.
However, it is not developed for UAV dodging because it requires high computational resources, making real-time onboard perception infeasible.
Therefore, our work aims to develop a solution that enables long-range detection, exhibits strong generalization capabilities, and effectively avoids the projectile using low-cost devices.

\vspace{-0.25cm}
\subsection{Human Pose Estimation in Intention-Aware Robotics}
\label{subsec:system_plant}
With recent advancements in HPE, including enhanced accuracy, multi-person tracking, higher frame rates, and improved generalization~\cite{2020CoRR_BlazePose_Bazarevsky, 2021TPAMI_OpenPose_Cao, 2023TPAMI_AlphaPose_Fang}, pretrained HPE models are increasingly integrated into downstream robotics applications. As a result, HPE has attracted considerable attention for improving autonomous navigation in scenes involving human movement. It provides a vital cue for understanding motion and predicting intentions, particularly in improving SLAM in dynamic scenes~\cite{2022ICRA_AirDOS_Qiu}, forecasting 2D human trajectories in crowded settings using a transform-based framework~\cite{2023RAL_HST_Salzmann}, avoiding pedestrian dynamics~\cite{xu2025navrl}, and enabling safe human tracking with UAV~\cite{2024IROS_IntentionAwarePlanner_Ren}. Although these methods primarily focus on slow-moving pedestrian scenarios, the idea of integrating human intent into robotic systems has inspired our work. Meanwhile,~\cite{zhu2023motionbert} learns human motion representations and enables semantic-level intent understanding, but it cannot operate in real time. Building on this, we propose leveraging HPE to enable UAVs to timely dodge human aggressive actions, enhancing the robustness against sudden fast projectiles.

\vspace{-0.25cm}

\section{Methodology}
\label{sec:predictor_design}

\begin{figure*}[t]
\vspace*{-0.2cm}
\centering
\includegraphics[width=0.95\textwidth]{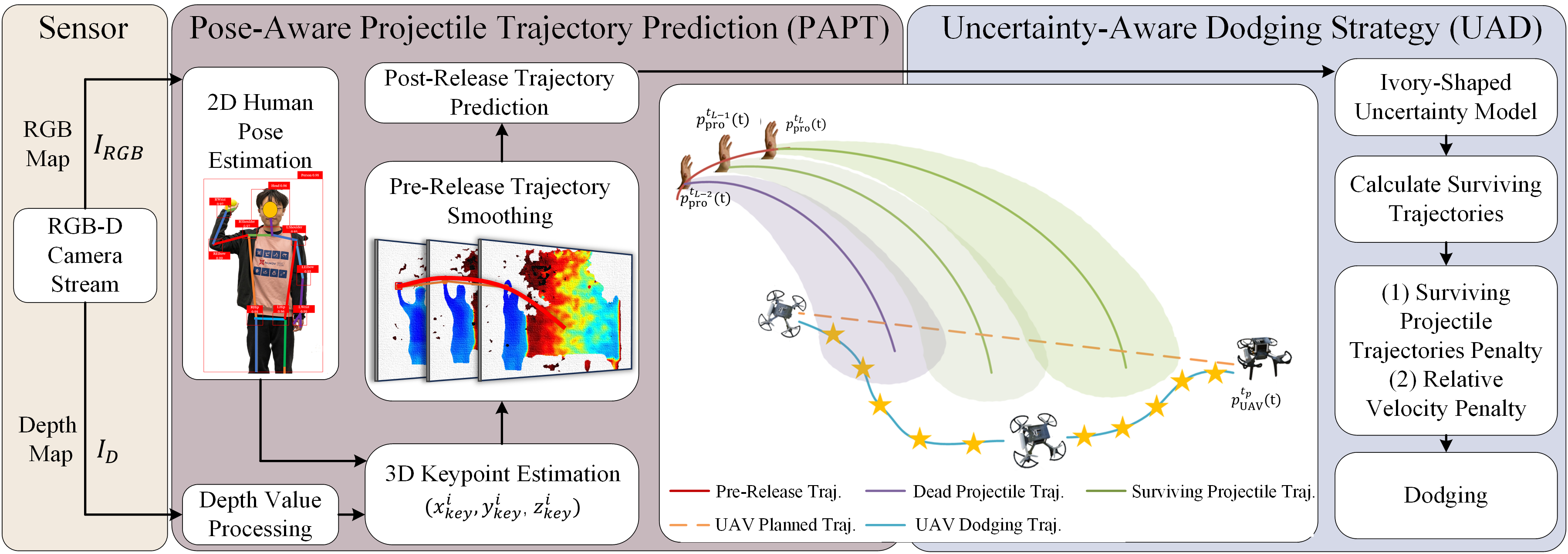}
\vspace{-0.2cm}  
\caption{\textbf{System overview.}
The system architecture for UAV dodging of human-initiated projectile attacks consists of two main modules: PAPT and UAD. PAPT takes RGB-D camera streams as input, processes RGB-D camera streams through 2D HPE and depth value processing to track 3D body keypoints associated with projectile throwing, and calculates pre- and post-release trajectory. Using initial projectile trajectory predictions, the UAD employs an ivory-shaped uncertainty model and identifies surviving trajectories and regions for assessing UAV collision risk. It then incorporates penalties from the proximity and relative velocity of surviving projectile trajectories to generate a safe dodging trajectory.
\label{fig:overview}}
\vspace{-0.3cm}
\end{figure*}

This work presents a novel framework that enables the UAV to dodge intentional human-initiated attacks. The proposed approach comprises two core modules: PAPT and UAD.

\vspace{-6pt}
\subsection{Pose-Aware Projectile Trajectory Prediction (PAPT)}
\label{subsec:PAPT}

The PAPT framework is a real-time system that integrates HPE with projectile motion modeling to track human joint movements and predict the trajectory of the thrown projectile. Specifically, it combines modules for 3D Keypoint Estimation and Projectile Trajectory Prediction, enabling the UAV to assess and respond to potential threats effectively. 

\subsubsection{3D Keypoint Estimation}

PAPT leverages synchronized RGB-depth image pairs $(\bm{I}_{\text{RGB}}, \bm{I}_{\text{D}})$, where $\bm{I}_{\text{RGB}}$ denotes the RGB map and $\bm{I}_{\text{D}}$ the depth map, to estimate 3D human joint positions in real time to anticipate projectile trajectory. The system first extracts 2D pose positions $(u_{\text{key}}^{i}, v_{\text{key}}^{i})$ for each critical joint (e.g., wrists) in each frame $i$ from $\bm{I}_{\text{RGB}}$ using RTMPose~\cite{2023CoRR_RTMPose_Jiang}, then maps these positions to the corresponding locations in the depth map $\bm{I}_{\text{D}}$ to obtain the depth values $d_{\text{key}}^{i}$. To mitigate noise inherent in depth sensing, spatial and temporal consistency are applied to ensure the reliability of $d_{\text{key}}^{i}$. Spatially, we compute the average depth within a small window centered at $(u_{\text{key}}^{i}, v_{\text{key}}^{i})$ after removing outliers, discarding the frame if the result remains unreliable. Temporally, we compare the current frame’s depth with the previous valid frame, discarding frames where the difference exceeds a predefined threshold. 

Thus, the 3D coordinates $(x_{\text{key}}^{i},y_{\text{key}}^{i},z_{\text{key}}^{i})$ in the world frame are obtained using the camera intrinsic matrix $\mathbf{K}$ and the transformation matrix $\mathbf{T}_{\mathcal{WC}}$ from camera to world frame:
\begin{equation}
\begin{bmatrix}
x_{\text{key}}^i \\
y_{\text{key}}^i \\
z_{\text{key}}^i \\
1
\end{bmatrix}
=
\mathbf{T}_{\mathcal{WC}} \cdot
\begin{bmatrix}
d_{\text{key}}^i \cdot \mathbf{K}^{-1}
\begin{bmatrix}
u_{\text{key}}^i \\
v_{\text{key}}^i \\
1
\end{bmatrix}
\\
1
\end{bmatrix}
\label{eq:xyz_world_from_pixel}
\end{equation}

\subsubsection{Projectile Trajectory Prediction} 
The PAPT framework predicts projectile trajectories by modeling the dynamics of human keypoints in two phases: Pre-Release Trajectory Smoothing and Post-Release Trajectory Prediction.

\paragraph{Pre-Release Trajectory Smoothing}
We utilize cubic spline regression to fit the 3D keypoint positions $(x_{\text{key}}^{i},y_{\text{key}}^{i},z_{\text{key}}^{i})$, thereby obtaining the pre-release trajectory. To reduce computational overhead, only the most recent $N$ valid frames are considered. To avoid zig-zag trajectory, we fit cubic splines independently along each axis: $\left(t_{\text{key}}^{i}, x_{\text{key}}^{i}\right)$, $\left(t_{\text{key}}^{i}, y_{\text{key}}^{i}\right)$, and $\left(t_{\text{key}}^{i}, z_{\text{key}}^{i}\right)$. The smoothing factor is empirically selected. Here, $t^{i}$ denotes the timestamp of frame $i$ for $i = n-N+1, \ldots, n$, where \(n\) is the current frame number. Consequently, the smoothed keypoint trajectories along each axis are obtained as $\bm{x}_{\text{key}}(t)$, $\bm{y}_{\text{key}}(t)$, and $\bm{z}_{\text{key}}(t)$. The position of the keypoint at time $t$ along the pre-release trajectory is then represented as $\bm{p}_{\text{key}}(t) = \left(\bm{x}_{\text{key}}(t), \bm{y}_{\text{key}}(t), \bm{z}_{\text{key}}(t)\right)^\top$. The corresponding velocity $\bm{v}_{\text{key}}(t)$ and acceleration $\bm{a}_{\text{key}}(t)$ are derived from the spline-interpolated position $\bm{p}_{\text{key}}(t)$ as its first- and second-order time derivatives, respectively.

\paragraph{Post-Release Trajectory Prediction}
Given the inherent unpredictability of human decision-making, which results in stochastic timing of throwing, a conservative approach is adopted to predict the post-release trajectory. Specifically, we consider any time instant $t_L$ as a potential release time if it satisfies the conditions that the magnitude of the pre-release keypoint acceleration  $\|\bm{a}_\text{key}(t_L)\|$ exceeds a predefined threshold $\theta_a$, and directions of the pre-release velocity and acceleration are consistent (verified by checking if the dot product $\bm{a}_{\text{key}}(t_L) \cdot \bm{v}_{\text{key}}(t_L)$ is positive).
We use the corresponding position $\bm{p}_{\text{key}}(t_L)$ and velocity $\bm{v}_{\text{key}}(t_L)$ as initial conditions to predict the projectile's velocity $\bm{v}_{\text{pro}}^{t_L}(t)$ and position $\bm{p}_{\text{pro}}^{t_L}(t)$ for a potential release at $t_L$, modeled as:
\begin{equation}
\begin{aligned}
\bm{v}_{\text{pro}}^{t_L}(t) &= \bm{v}_{\text{key}}(t_L) + \bm{g} t, \\
\bm{p}_{\text{pro}}^{t_L}(t) &= \bm{p}_{\text{key}}(t_L)+\bm{v}_{\text{key}}(t_L) t+\frac{1}{2}\bm{g}t^2.
\end{aligned}
\end{equation}
where $\bm{g} = [0, 0, -g]^\top$ is the gravitational acceleration.

\vspace{-5pt}
\subsection{Uncertainty-Aware Dodging Strategy (UAD)} 

Beyond the uncertainty in the release timing, the predicted trajectory also exhibits uncertainty. We propose a UAD module that enables the UAV to dodge the approaching projectile by simultaneously accounting for both factors.

Accurate prediction of projectile trajectory is essential for UAVs to dodge threats in human-initiated attacks. However, estimation errors in initial conditions, nonlinear aerodynamic effects, and environmental noise hinder high-precision prediction using ideal parabolic models~\cite{2015ECMR_Towards_Mueggler}. To address this challenge, we introduce an ivory-shaped uncertainty model that augments an ideal parabolic trajectory with a temporally expanding uncertainty envelope.
\subsubsection{Ivory-Shaped Uncertainty Model}

Multiple sources of uncertainty lead to cumulative errors in trajectory prediction over time. These errors manifest as an expanding uncertainty region around the predicted trajectory. To model this uncertainty, we represent the possible positions at each time $t$ as a three-dimensional ball with uncertainty radius $R(t)$. Nonlinear aerodynamics contribute a quadratic term $\alpha t^2$ to the uncertainty of position, while initial velocity and position errors add linear $\beta t$ and constant $\gamma$ terms. Thus, the uncertainty radius is modeled as:
\begin{equation}
R(t) = \alpha t^2 + \beta t + \gamma, \quad t \in [0, T_s^{t_L}].
\end{equation}
where $\alpha$, $\beta$, and $\gamma$ are empirically determined parameters, and $T_s^{t_L}$ denotes the survival duration of the post-release trajectory, which is the time from the projectile's release instant to its landing. A post-release trajectory is classified as surviving if $t < t_L + T_s^{t_L}$ and as dead (i.e., non-threatening) otherwise. For a post-release trajectory released at time $t_L$, the possible position distribution at future times $t \in [0, T_s^{t_L}]$ is given by:
\begin{equation}
\mathcal{P}^{t_L}(t) = \left\{ \bm{p} \in \mathbb{R}^3 \mid \left\| \bm{p} - \bm{p}_{\text{pro}}^{t_L}(t) \right\|^2 \leq R^2(t) \right\}.
\end{equation}
This formulation defines an ivory-shaped uncertainty region, signaling collision risk upon intersection with the UAV's position or planned trajectory.
\begin{figure}[htbp]
\vspace*{-0.2cm}
	\centering
	\includegraphics[width=0.40\textwidth]{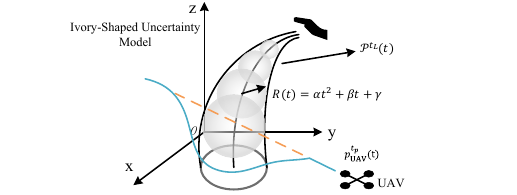}
    \vspace{-0.1cm}
	\caption{\textbf{Ivory-shaped uncertainty model:} 
    The prediction error in the projectile's trajectory increases over time. Based on this, we construct an ivory-shaped uncertainty region around the predicted trajectory. A collision risk is deemed present when the UAV’s current position or planned trajectory intersects with this region.
	\label{fig:Ivoryshaped}}
\vspace*{-0.2cm}
\end{figure}

\subsubsection{UAV Trajectory Optimization}

The UAV’s trajectory is parameterized using the MINCO polynomial class~\cite{2022TRO_GCOPTER_MINCO_Wang}:
\begin{equation}
\begin{aligned}
\mathfrak{T}_{\text{MINCO}} = &\big\{ \bm{p}_{\text{UAV}}^{t_p}(t) : [0, T_u] \mapsto \mathbb{R}^m \mid \bm{c} = \mathcal{M}(\bm{q}, \bm{T}), \\
&\bm{q} \in \mathbb{R}^{m(M-1)}, \bm{T} = [T_1, \ldots, T_M]^T \in \mathbb{R}_{>0}^M \big\}
\end{aligned}
\label{eq:minco}
\end{equation}

where $\bm{p}_{\text{UAV}}^{t_p}(t)$ denotes an $m$-dimensional UAV trajectory planned at time $t_p$, composed of $M$ quintic polynomial segments, and $T_u$ is the total trajectory duration. The trajectory is compactly parameterized by the intermediate waypoints $\bm{q}$ and segment durations $\bm{T}$, while the polynomial coefficients $\bm{c} =(\bm{c}_{1}^\top, \ldots, \bm{c}_{M}^\top)^\top$ are computed via a linear mapping $\mathcal{M}(\bm{q}, \bm{T})$. Given this parameterization, any trajectory cost $\mathcal{J}(\bm{q}, \bm{T})= \mathcal{F}(\bm{c},\bm{T})= \mathcal{F}(\mathcal{M}(\bm{q},\bm{T}), \bm{T})$ can be evaluated and optimized over $\bm{q}$ and $\bm{T}$, where $\mathcal{F}$ is any user-defined objective. Thus, the gradients $\frac{\partial \mathcal{J}}{\partial \bm{q}}$ and $\frac{\partial\mathcal{J}}{\partial \bm{T}}$ are propagated from the gradients $\frac{\partial \mathcal{F}}{\partial\bm{c}}$ and $\frac{\partial \mathcal{F}}{\partial \bm{T}}$ respectively, to optimize the trajectory cost $\mathcal{J}$. We formulate the trajectory generation as an unconstrained optimization problem. The optimization objective is given by:

\begin{equation}
\begin{aligned}
&\min_{\bm{c}, \bm{T}}\;
(\omega_s \mathcal{J}_s + \omega_o \mathcal{J}_o + \omega_t \mathcal{J}_t + \omega_f \mathcal{J}_f)
+ \omega_d \mathcal{J}_d + \omega_v \mathcal{J}_v, \\
&\omega_s{=}\omega_o{=}\omega_f{=}10.0,\; \omega_t{=}0.5,\; 
\omega_d{=}20.0,\; \omega_v{=}5.8,\; T{=}4.0
\end{aligned}
\label{eq:optimization}
\end{equation}

where $\mathcal{J}_s$, $\mathcal{J}_o$, $\mathcal{J}_t$, and $\mathcal{J}_f$ respectively denote the smoothness, static obstacle, time, and feasibility costs. The constraints and gradient computations follow the GCOPTER framework~\cite{2022TRO_GCOPTER_MINCO_Wang}. To handle projectile threats, we introduce two novel penalty terms: $\mathcal{J}_d$, which discourages the UAV trajectory from approaching predicted projectile trajectories, and $\mathcal{J}_v$, which promotes velocity directions favorable for dodging. The corresponding weights $\omega_d, \omega_v > 0$ control their relative importance. The proposed cost formulation is general and platform-independent, allowing the same set of weights to be transferred across different UAV platforms with only minor re-tuning for variations in agility and dynamic limits.

\begin{figure*}[t]
\vspace*{-0.35cm}
	\centering
	\includegraphics[width=0.95\textwidth]{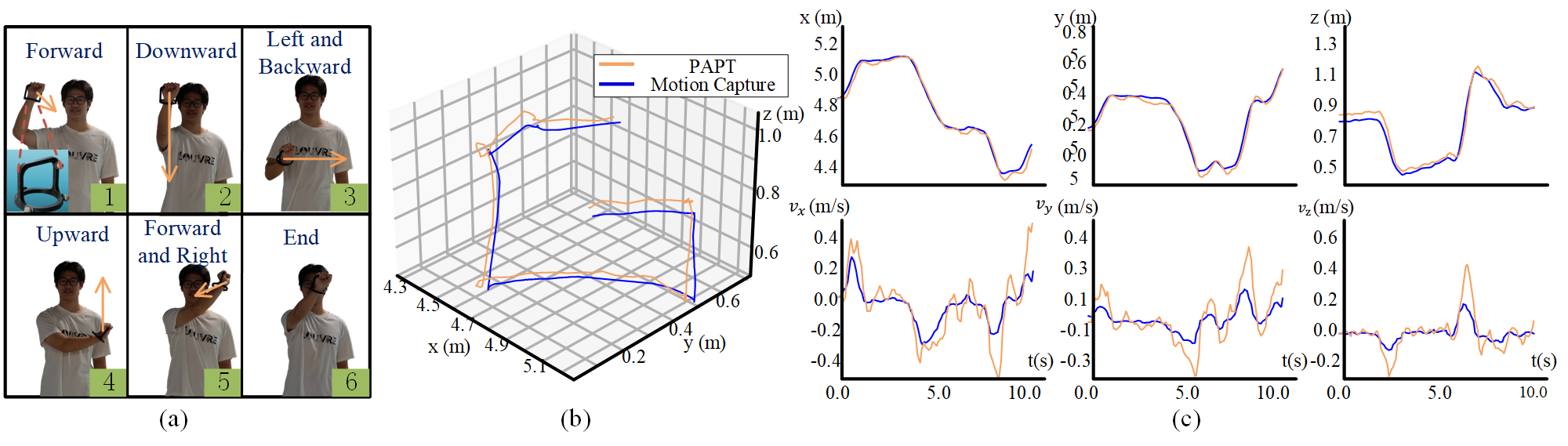}
    \vspace{-0.3cm}
	\caption{(a) A participant performs arm motions while wearing a 3D-printed device equipped with mocap markers.
    (b) Comparison of PAPT-estimated 3D trajectory with ground truth from the mocap.
    (c) Positional and velocity differences between the estimated and ground truth along the x-, y-, and z-axes.
	\label{fig:accuracy}}
\vspace*{-0.35cm}
\end{figure*}

\paragraph{Surviving Projectile Trajectories Penalty}

Dodging is triggered if the UAV's position or planned trajectory intersects an ivory-shaped uncertainty region $\mathcal{P}^{t_l}(t)$ associated with a projectile released at $t_l$. Accurately predicting projectile release timing poses a key challenge in trajectory planning. Relying solely on the current observed trajectory of a projectile can lead to significant limitations. Specifically, planning algorithms may erroneously interpret the current trajectory as the actual trajectory, thus overlooking latent risks embedded in historical trajectory data. Such misjudgments can cause the planner to optimize the trajectory toward regions perceived as collision-free, which, in reality, may coincide with the projectile's actual trajectory, thereby incurring collision hazards. To address this issue, we consider all surviving post-release trajectories at the current planning time $t_p$, represented as:\begin{equation}
\begin{aligned}
& \bm{p}_{\text{pro}}^{t_l}(t), \quad l = 1, 2, \ldots, L, \quad t \in [0, T_u].
\end{aligned}
\end{equation}
subject to $t_p > t_l$ and $t_l + T_s^{t_l} > t_p$ for all $l$, where each such trajectory is a potential candidate for the actual trajectory, and $L$ is the number of surviving projectile trajectories. 
To address this uncertainty, we penalize the proximity between the UAV's future sampling points and predicted projectile positions:

{\small
\begin{equation}
\begin{aligned}
& \mathcal{J}_d = \sum_{m,k,l}\Big[ \max\Big(0,\ R(\tau_{m,k} + \Delta t_{p,l}) + R_s - ||\bm{d}_{m,k,l}|| \Big) \Big]^2,\\
& \bm{d}_{m,k,l} = \bm{p}_{\text{UAV}}^{t_p}(\tau_{m,k}) - \bm{p}_{\text{pro}}^{t_l}(\tau_{m,k} + \Delta t_{p,l}), \Delta t_{p,l} =  t_p - t_l,\\
& \tau_{m,k} = \sum_{j=0}^{m-1} T_j + \frac{k}{K} T_m, m=1, \ldots, M, k=0,\ldots, K-1,
\end{aligned}
\end{equation}
}
where $\sum_{m,k,l} = \sum_{m=1}^M \sum_{k=0}^{K-1} \sum_{l=1}^L$, $K$ represents the number of sampling points on each piece of the UAV's trajectory, $R_s$ denotes the UAV’s safety margin radius, and $\Delta t_{p, l}$ represents the time elapsed since the $l$-th projectile is released. The gradients of $\mathcal{J}_d$ w.r.t. $\bm{c}_i$ and $T_i$ can be derived by the chain rule:

{\small
\vspace{-0.2cm}
\begin{equation}
\begin{aligned}
\frac{\partial \mathcal{J}_d}{\partial \bm{q}} &= \sum_{m,k,l} \frac{\partial \mathcal{J}_d}{\partial \bm{p}_{\text{UAV}}^{t_p}} \frac{\partial \bm{p}_{\text{UAV}}^{t_p}}{\partial \bm{c}_m}, \\
\frac{\partial \mathcal{J}_d}{\partial T_i} &= \sum_{m,k,l} \frac{\partial \mathcal{J}_d}{\partial \bm{p}_{\text{UAV}}^{t_p}}\left( \frac{\partial \bm{p}_{\text{UAV}}^{t_p}}{\partial T_i} + \frac{\partial \bm{p}_{\text{UAV}}^{t_p}}{\partial \bm{c}_m} \frac{\partial {c}_m}{\partial T_i} \right)
\end{aligned}
\label{eq:gradients}
\end{equation}
}

\paragraph{Relative Velocity Penalty}
To mitigate collision risk, we penalize the component of the relative velocity between the UAV and projectile onto the relative position vector, encouraging adjustments in speed magnitude or lateral maneuvers.
\begin{equation}
\begin{aligned}
\mathcal{J}_v &= \sum_{m,k,l}\left( \frac{\bm{v}_{\text{rel},m,k,l} \cdot \bm{d}_{m,k,l}}{\|\bm{d}_{m,k,l}\| + \epsilon} \right)^2, \\
\bm{v}_{\text{rel},m,k, l} &= \bm{v}_{\text{UAV}}^{t_p}(\tau_{m,k}) - \bm{v}_{\text{pro}}^{t_l}(\tau_{m,k} + \Delta t_{p,l}) \\
\end{aligned}
\label{eq:cost_function}
\end{equation}
where \(\bm{v}_{\text{pro}}^{t_l}(t)\) and \(\bm{v}_{\text{UAV}}^{t_p}(t)\) are the velocities at time \(t\) of the projectile released at \(t_l\) and the UAV whose trajectory is planned at \(t_p\), and \(\epsilon > 0\) ensures numerical stability. The gradient computation follows the method outlined in \eqref{eq:gradients}: 
{\small
\begin{equation}
\begin{aligned}
\frac{\partial \mathcal{J}_v}{\partial \bm{q}} &= \sum_{m,k,l} \left( \frac{\partial \mathcal{J}_v}{\partial \bm{p}_{\text{UAV}}^{t_p}} + \frac{\partial \mathcal{J}_v}{\partial \bm{v}_{\text{UAV}}^{t_p}} \right) \frac{\partial \bm{p}_{\text{UAV}}^{t_p}}{\partial \bm{c}_m} , \\
\frac{\partial \mathcal{J}_v}{\partial T_i} &= \sum_{m,k,l} \left( \frac{\partial \mathcal{J}_v}{\partial \bm{p}_{\text{UAV}}^{t_p}} + \frac{\partial \mathcal{J}_v}{\partial \bm{v}_{\text{UAV}}^{t_p}}  \right) 
\left( \frac{\partial \bm{p}_{\text{UAV}}^{t_p}}{\partial T_i} + \frac{\partial \bm{p}_{\text{UAV}}^{t_p}}{\partial \bm{c}_m}\frac{\partial \bm{c}_m}{\partial T_i} \right)
\end{aligned}
\label{eq:gradients2}
\end{equation}
}


\section{EXPERIMENT}
\label{sec:quad_experiment_verification}

\subsection{Hardware Platform}

We evaluate our method on a $1.35~\si{kg}$ custom quadrotor ($240{\times}240{\times}105~\si{mm}$) integrating an Intel RealSense D455 RGB-D camera, a Minisforum EM780 (AMD Ryzen 7 7840U CPU) computer, and an NxtPX4v2 flight controller.

\vspace{-0.2cm}

\subsection{Evaluation of the PAPT Module}
\subsubsection{3D Pose Keypoint Tracking Accuracy} 
To evaluate the accuracy of our PAPT algorithm, we compare its keypoint position estimates against ground truth (GT) data obtained from a Nokov motion capture (mocap) system. A human participant performs wrist movements with a reflective marker rigidly attached via a custom 3D-printed fixture, which serves as the GT reference (Fig.~\ref{fig:accuracy} (a)). The subject executes unconstrained wrist motions, and we quantify the tracking errors between PAPT estimates and the mocap GT.

The 3D keypoint tracking performance of PAPT closely aligns with mocap GT across all spatial dimensions. For position estimates, PAPT achieves a root mean squared error (RMSE) of 0.037 m and a maximum error (MaxE) of 0.0876 m. For velocity estimates, it yields an RMSE of 0.154 m/s and a MaxE of 0.5257 m/s. Specifically, the mean positional errors along the $x-$ and $y-$axes achieve sub-centimeter accuracy ($ < 0.01 m$), while the $z-$axis exhibits a slightly higher bias of approximately 0.021 m. Mean velocity errors are slightly elevated at instances of abrupt motion changes, approximately 0.192 m/s. Despite these deviations, errors remain within acceptable limits for real-time applications.

\subsubsection{Detection Distance and Time Consumption}
\label{subsubsec:Analysis_TimeConsumption}

To evaluate the PAPT's perception capability, we reproduce the algorithm from~\cite{2022RAL_FastDodging_Lu} on our platform and conduct 20 trials. Two key indicators are statistically assessed: (1) the maximum effective detection distance, defined as the farthest range at which the 3D position of the target can be perceived (human keypoints in our case, vs. tennis ball in~\cite{2022RAL_FastDodging_Lu}), and (2) the perception latency, defined as the time from target detection to its trajectory generation. The results are summarized in Table~\ref{tab:perception_method_comparison}.

\begin{table}[h]
\vspace*{-0.3cm}
\centering
\caption{Comparison of Detection Distance and Latency}
\label{tab:perception_method_comparison}
\renewcommand{\arraystretch}{0.95}  
\setlength{\tabcolsep}{3pt}        
\begin{threeparttable}
\begin{tabularx}{\columnwidth}{l *{2}{>{\centering\arraybackslash}X}}
\toprule
 & Lu et al.~\cite{2022RAL_FastDodging_Lu} & Ours \\
\midrule
Effective Distance (m) & 4.2\tnote{1} / 3.9\(^*\) & \textbf{6.0}\tnote{1} \\
Average Time Cost (ms) & 320\tnote{1} / 29\(^*\) & \textbf{26.4}\tnote{1} \\
\bottomrule
\end{tabularx}
\begin{tablenotes}
\footnotesize
\item[\tnote{1}]CPU results reproduced on our platform; \(^*\) GPU results reported in~\cite{2022RAL_FastDodging_Lu}.
\end{tablenotes}
\end{threeparttable}
\vspace*{-0.3cm}
\end{table}

\begin{figure*}[t]
\vspace*{-0.3cm}
	\centering
	\includegraphics[width=0.95\textwidth]{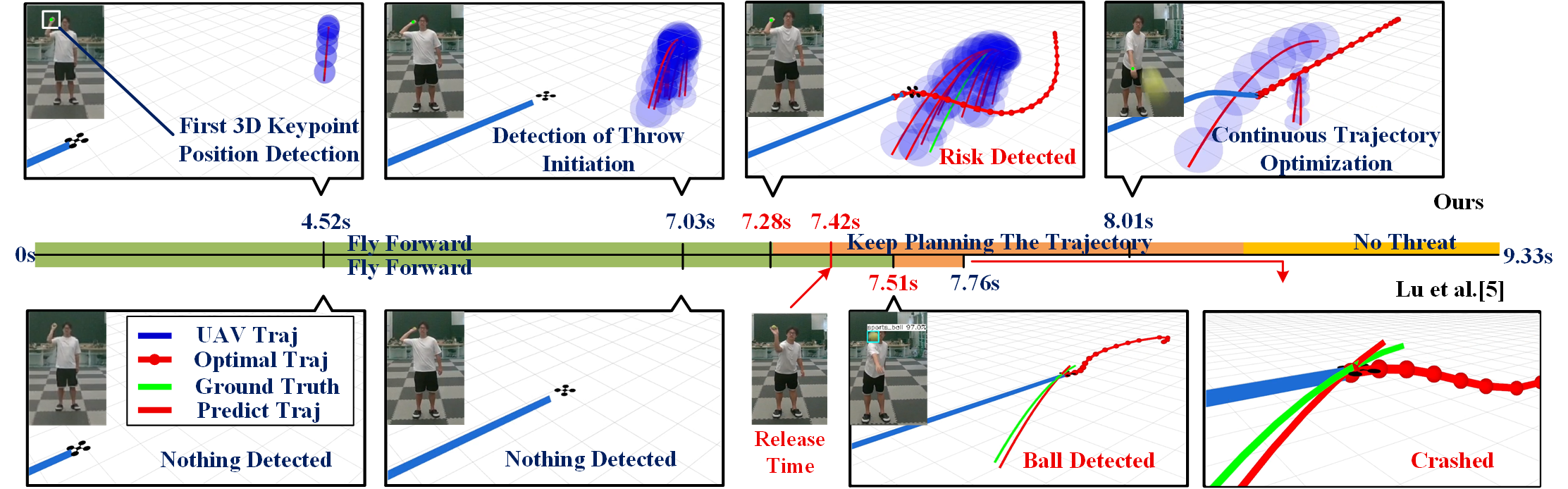}
    \vspace{-0.1cm}  
	\caption{Timeline analysis illustrating system performances for attack dodging within the same simulated environment. The figure presents a horizontal time axis with processing operations at distinct time nodes for two systems.
	\label{fig:timeline}}
\vspace*{-0.3cm}
\end{figure*}

Our method achieves an effective detection distance of \SI{6}{m} and an average latency of \SI{26.4}{ms}, outperforming the approach in~\cite{2022RAL_FastDodging_Lu} in terms of detection range and responsiveness. Unlike direct object detection, PAPT infers projectile positions by recognizing the entire human skeletal structure, thereby eliminating the need for the projectile to be visible or to meet a minimum size threshold. This approach significantly extends the effective detection range. Our algorithm accurately identifies the 2D positions of keypoints at \SI{6}{m}, a limit imposed by the hardware constraints of the camera's depth sensing capabilities, suggesting that \SI{6}{m} is not the true upper bound of our method's capability. Notably, our PAPT algorithm is implemented on a CPU platform, achieving an average latency lower than that of~\cite{2022RAL_FastDodging_Lu}, which fails to enable real-time detection when tested on a CPU. This highlights the versatility of our method, enabling effective detection on lower-cost platforms.

\subsubsection{Evaluation under diverse lighting}
\label{subsubsec:diverse lighting}

The PAPT algorithm is further tested under various lighting conditions, including dim-light, planar illumination, and different spotlight settings. Across all conditions, the wrist position RMSE remains below 0.09 m, indicating stable perception performance. Only under extremely bright illumination (40W spotlight) do short-term detection failures occur due to overexposure and the resulting loss of visual features, after which the tracking quickly recovered, demonstrating the robustness of the perception module.

\vspace{-0.32cm}

\subsection{Evaluation of the UAD Module}
\label{Evaluation of the Uncertainty Dodging Strategy}

To validate the effectiveness of our UAD module, we conduct an ablation study in simulation. A collision is deemed to occur when the minimum distance $d_{\min}$ between the UAV and the actual projectile is less than $0.4~\si{m}$. The experimental procedure is as follows: (1) The UAV is controlled to fly in a real environment while collecting RGB-D images and trajectory data; a tennis ball is thrown toward the UAV at the final moment. (2) The recorded data are then imported into the simulation environment to test variations of our method. To ensure accurate synchronization between RGB-D information and positional data, the UAV follows the pre-recorded trajectory while processing the RGB-D images for perception. Once a potential threat is detected, the system immediately switches to its projectile dodging strategy and plans a trajectory to reach the predefined goal. Ours-NoTemporal method solely considers the single projectile trajectory predicted at the current timestep. Ours-NoSpatial incorporates temporal evolution but assumes perfect prediction accuracy, ignoring the spatial uncertainty. We perform 30 trials across diverse scenarios, recording the mean minimum distance and SR.

\begin{table}[h]
\vspace*{-0.28cm}
    \centering
    \caption{Performance Comparison of Projectile Dodging Strategies}
    \vspace{-0.1cm}
    \label{tab:Evaluationofback} 
    \begin{tabular}{lccc}
    \toprule
     & Ours-NoTemporal & Ours-NoSpatial  & Ours\\
    \midrule
    Average $d_{\text{min}}$ $(m)$ & 0.38 & 0.51 & \textbf{0.85}\\
    SR(\%) & 43.33 & 86.67 & \textbf{100} \\
    \bottomrule
    \end{tabular}
\vspace*{-0.2cm}
\end{table}

As shown in Table~\ref{tab:Evaluationofback}, our method achieves the best performance in terms of both average minimum distance $d_{\text{min}}$ and dodging SR. The Ours-NoTemporal variant suffers the most, as it dodges non-existent projectile trajectories for most of the time, resulting in collisions with the actual projectile trajectory. This approach relies heavily on accurately predicting the projectile's release time—a task that is exceedingly challenging in practice. Ours-NoSpatial accounts for temporal uncertainty but neglects spatial uncertainty, assuming that a set of predicted trajectories includes the actual projectile trajectory. This leads to a degraded SR because prediction errors cause inadequate dodging maneuvers, resulting in collisions. In contrast, our method considers both temporal and spatial uncertainties, enabling the UAV to anticipate and avoid threats by reacting to the broader, dynamically evolving risk region.

\begin{figure*}[htbp]
\vspace*{-0.4cm}
	\centering
	\includegraphics[width=0.96\textwidth]{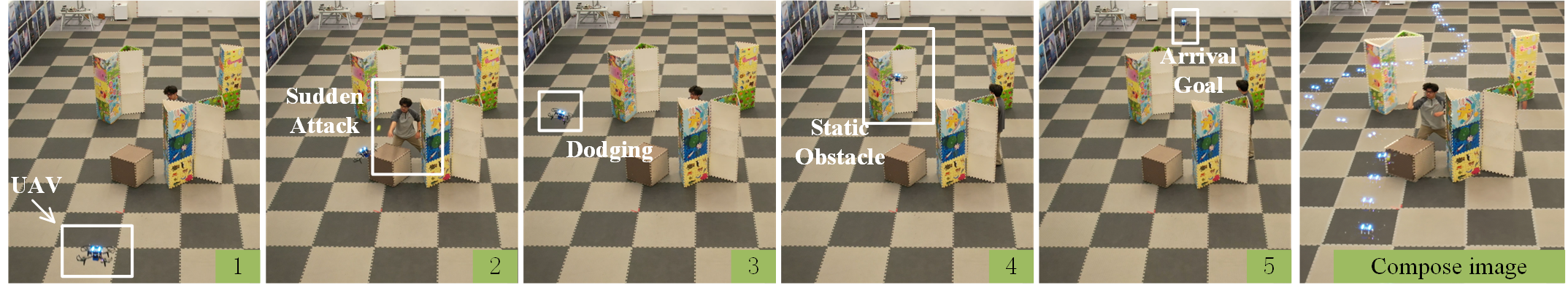}
    \vspace{-0.15cm}
	\caption{Sequence of a UAV detecting a sudden attack, avoiding the attack and static obstacles, and reaching the target during flight.
	\label{fig:realflight}}
\vspace*{-0.2cm}
\end{figure*}

\begin{table*}[htbp]
\vspace*{-0.2cm}
\centering
\renewcommand{\arraystretch}{0.95}   
\setlength{\tabcolsep}{2.5pt}        
\scriptsize                          
\caption{Dodging SR Across Various Conditions While Hovering}
\label{tab:success_rate_extended}
\begin{tabularx}{\textwidth}{@{}p{0.9cm}p{1.6cm}*{16}{>{\centering\arraybackslash}X}@{}}
\toprule
\multicolumn{2}{@{}l}{} & \multicolumn{16}{c}{\textbf{Success Rate (\%)}} \\
\cmidrule(lr){1-2} \cmidrule(lr){3-18}
\multicolumn{2}{l}{\textbf{Attacker-to-UAV Dist.(m)}} &
\multicolumn{4}{c}{\textbf{3}} &
\multicolumn{4}{c}{\textbf{4}} &
\multicolumn{4}{c}{\textbf{5}} &
\multicolumn{4}{c}{\textbf{6}} \\
\cmidrule(lr){1-2}
\cmidrule(lr){3-6} \cmidrule(lr){7-10}
\cmidrule(lr){11-14} \cmidrule(lr){15-18}

\multicolumn{2}{l}{\textbf{Proj. Speed Group}} &
Low & Med & High & Ext &
Low & Med & High & Ext &
Low & Med & High & Ext &
Low & Med & High & Ext \\

\multicolumn{2}{l}{\textbf{\shortstack{Avg. Proj. Speed (m/s)}}} &
3.16 & 4.08 & 5.27 & 6.10 &
4.83 & 6.19 & 7.30 & 8.27 &
6.45 & 7.42 & 9.01 & 10.18 &
8.92 & 10.42 & 11.96 & 13.29 \\

\cmidrule(lr){1-2} \cmidrule(lr){3-18}
\multirow{3}{*}{\textbf{\shortstack{Attack\\Angle}}}
& \multicolumn{1}{r}{\textbf{$30^{\circ}$}}
& 100 & 100 & 85.7 & 71.4
& 100 & 100 & 100 & 85.7
& 100 & 85.7 & 100 & 100
& 100 & 100 & 71.4 & 71.4 \\
& \multicolumn{1}{r}{\textbf{$0^{\circ}$}}
& 100 & 100 & 85.7 & 100
& 100 & 100 & 85.7 & 71.4
& 100 & 100 & 100 & 85.7
& 85.7 & 100 & 85.7 & 85.7 \\
& \multicolumn{1}{r}{\textbf{$-30^{\circ}$}}
& 100 & 100 & 100  & 85.7
& 100 & 100 & 85.7 & 85.7
& 100 & 100 & 71.4 & 85.7
& 100 & 85.7 & 85.7 & 71.4 \\

\cmidrule(lr){1-2} \cmidrule(lr){3-18}
\multicolumn{2}{l}{\textbf{Overall}} &
100 & 100 & 90.5 & 81.0 &
100 & 100 & 90.5 & 81.0 &
100 & 95.2 & 90.5 & 90.5 &
95.2 & 95.2 & 81.0 & 76.2 \\
\bottomrule
\end{tabularx}
\label{tab:success_rate_optimized}
\vspace*{-0.35cm}
\end{table*}

\vspace{-0.3cm}

\subsection{Evaluation of the Overall System}
\label{Evaluation of the overall system}

To demonstrate the superiority of our overall framework, we conduct comparative experiments against the baseline in~\cite{2022RAL_FastDodging_Lu}  within a simulated environment. We utilize the experimental setup described in Section~\ref{Evaluation of the Uncertainty Dodging Strategy}. Both systems adopt their own perception modules for perception and switch to their respective dodging strategies upon detecting a potential attack.

Fig.~\ref{fig:timeline} illustrates a timeline for a representative dodging scenario, where UAVs initially follow the same pre-recorded trajectory in both systems. At \SI{4.52}{s}, our method first detects the 3D positions of human keypoints, \SI{2.99}{s} earlier than the baseline’s initial object detection~\cite{2022RAL_FastDodging_Lu}. Our method detects the throw initiation at \SI{7.03}{s}, and then predicts a potential collision risk at \SI{7.28}{s} via HPE and commences trajectory planning. The projectile is released at \SI{7.42}{s}, by which time planning has already been underway for \SI{0.14}{s}. Consequently, the UAV dodges the uncertainty region that encompassing the actual projectile trajectory. As the set of surviving projectile trajectories decreases, the UAV continuously refines its trajectory to reach the goal efficiently. In contrast, the baseline~\cite{2022RAL_FastDodging_Lu} detects the ball only at \SI{7.51}{s}, when it is in mid-air, affording insufficient planning time and resulting in a collision at \SI{7.76}{s}. This \SI{0.23}{s} detection advantage enables our successful dodging. Overall, we perform 30 trials across diverse scenarios and compute the corresponding SRs. Our approach yields a dodging SR of 96.67\%, markedly outperforming the baseline's 60\%, underscoring the efficacy against human-initiated threats.

\vspace{-0.3cm}

\subsection{Real-World Deployment and Evaluation}

We evaluate the dodging capabilities under two main real-world scenarios. In the first, a hovering UAV faces a projectile thrown by an attacker, as shown in Fig.~\ref{fig:cover}. In the second, the UAV navigates toward a goal while subjected to sudden human-initiated attacks, as shown in Fig.~\ref{fig:realflight}, where it performs an emergency lateral maneuver to dodge the projectile, reaches the target, and avoids collisions with static obstacles.

\subsubsection{Evaluation of Performance Limits}
This subsection evaluates the extreme dodging capabilities of our framework in real-world conditions. Because manually launched projectile speeds vary, directly measuring a single maximum dodge speed is infeasible. We therefore infer the system limit indirectly from dodging SRs. Specifically, for each attacker-to-UAV distance (3 m, 4 m, 5 m and 6 m) and attack angle (30$^{\circ}$, 0$^{\circ}$, and 30$^{\circ}$), we perform 21 trials with a spread of throwing intensities and group trials into four speed bands (Low, Medium, High, and Extreme, with each group represented by its respective average speed). We adopt the operational criterion that a speed band is considered to approach the UAV’s performance limit when the band exhibits two or more failures.

As shown in Table~\ref{tab:success_rate_optimized}, the results reveal that the UAV achieves high SRs in projectile dodging under low- and medium-speed conditions across all distances and angles, with an overall SR consistently at or above 95.2\%. At 3 m, the UAV reached a performance limit of approximately 5.27 m/s, as two failures occurred in the high-speed group. Analogously, the limits are inferred as 7.3 m/s at 4 m and 9.01 m/s at 5 m. At 6 m, the overall SR decreased, likely due to amplified prediction errors in projectile position and velocity as distance and speed escalate, leading to erroneous dodging decisions by the UAV. We further test higher projectile speeds in the Extreme group, where SR shows a slight decline. Nevertheless, even under the most challenging 6 m and 13.29 m/s condition, the UAV maintains an SR above 76.2\%, and achieves an overall SR of 82.4\% in the Extreme group, demonstrating robust performance for most human-initiated attack scenarios.

\subsubsection{Evaluation of System Robustness}
\label{System robustness}

To validate the robustness of our system, we conduct three sets of experiments. (1) A single participant performs throws using varied joint configurations (Right-hand Visible throw(RV), Right-hand Hidden throw(RH), Left-hand Visible throw(LV), Left-hand Hidden throw(LH)) and throwing motions (Downward, Upward, Right-to-Left, Left-to-Right) at a distance of 4 m from the hovering UAV, with throwing speeds around 6 m/s. Each unique combination of joint configuration and throwing motion is repeated 21 times. (2) Six participants (A–F), exhibiting diverse throwing styles and physical builds, throw projectiles of varying sizes and shapes (a tennis ball, a 3D-printed part, a bottle, a loopy toy, a football and a slipper) from the same distance and with speeds similar to those in the first experiment. Each configuration is also repeated 21 times. (3) A single participant stands approximately 4 m from the UAV and performs throwing motions with release speeds of about 6 m/s. Three occlusion conditions are tested: partial-arm occlusion, half-body occlusion, and full-body occlusion. Each condition is repeated 21 times to ensure statistical reliability.

\begin{figure*}[htbp]
\vspace*{-0.55cm}
	\centering
	\includegraphics[width=1.02\textwidth]{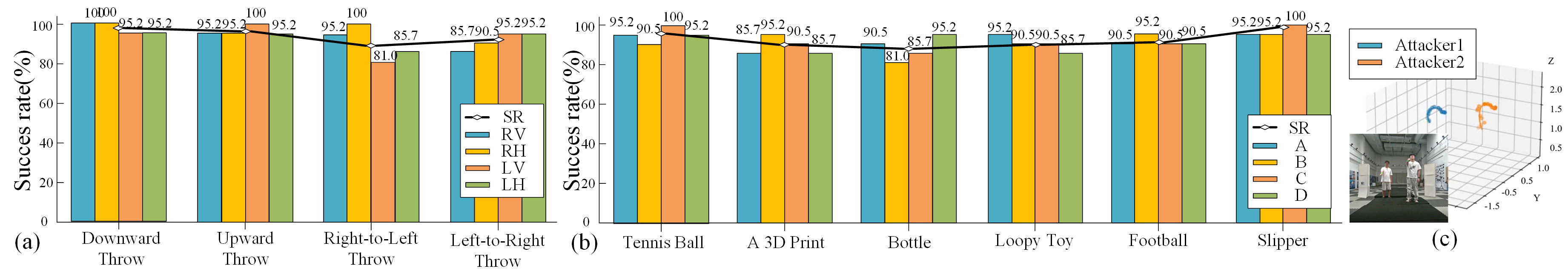}
    \vspace{-0.7cm}  
	\caption{(a) Success rates under varied throwing styles and joint configurations by a single participant. (b) Success rates with different projectiles thrown by four participants. (c) Detection and prediction of two simultaneous attacks.
	\label{fig:success_rate}}
\vspace*{-0.5cm}
\end{figure*}

As shown in Fig.~\ref{fig:success_rate}(a), the system exhibits consistent performance across diverse movement configurations, evidenced by a standard deviation (SD) of 5.45\% and a coefficient of variation (CV) of 5.8\% in the dodging SR for various joint positions and throwing motions. By leveraging human pose detection, the system effectively infers the projectile trajectory from global skeletal data, even in the presence of occlusions. Notably, hidden throws achieve higher SR compared to visible throws, attributable to the extended preparatory motions in concealed throws, which elongate the wind-up phase and provide the UAV with approximately 20 ms of additional reaction time. Throws using the non-dominant hand (e.g., left-handed right-to-left motion) exhibit reduced SR owing to wrist instability and excessive force, leading to inaccuracies in trajectory prediction. As illustrated in Fig.~\ref{fig:success_rate}(b), the second experimental set presents an SD of 4.60\% and a CV of 5.01\%, indicating minimal influence from variations in participants and projectiles. We also test dual-attacker simultaneous throws toward a hovering UAV, the system successfully identifies and tracks both attack trajectories and executes coordinated avoidance maneuvers, as shown in Fig.~\ref{fig:success_rate}(c).

\begin{table}[htbp]
\vspace*{-0.28cm}
  \centering
  \caption{Dodging SR under different occlusion conditions.}
  \label{tab:occlusion_SR}
  \renewcommand{\arraystretch}{1.0}
  \begin{tabular}{cccc}
    \toprule
    \textbf{Condition} & Partial-Arm & Half-Body & Full-Body \\
    \midrule
    \textbf{Success Rate (\%)} & 95.2 & 95.2 & 90.4 \\
    \bottomrule
  \end{tabular}
\vspace*{-0.22cm}
\end{table}

In occlusion experiments, the UAV achieves an average dodging SR of 93.6\%, demonstrating reliable performance even with incomplete visual information. The minor differences among occlusion conditions arise because human attackers generally follow a similar motion pattern, emerging from behind obstacles before releasing the projectile, which enables the perception module to extract sufficient cues for threat prediction. Overall, all these findings underscore the system's robustness and adaptability in varied scenarios.

\vspace{-0.2cm}

\section{CONCLUSION AND FUTURE WORK}
\label{sec:conclusion}

In this letter, we propose a novel real-time framework that enables UAVs to dodge small and fast projectiles using an RGB-D camera. The system integrates a Pose-Aware Projectile Trajectory Prediction module and an Uncertainty-Aware Dodging module, allowing the UAV to anticipate threats from human motion cues and execute agile maneuvers under uncertainty, and can be easily transferred across different UAV platforms due to its modular perception–planning architecture. Beyond this specific case, the framework contributes to enhancing human–UAV interaction safety, where UAVs interpret human motion cues to operate safely in dynamic and uncertain environments. Potential applications include sports broadcasting and aerial filming, where UAVs avoid incoming balls; security and patrolling missions, where they cope with sudden human-initiated threats; and crowded public events, where they predict local motion trends to navigate safely. Future work will pursue semantic understanding of human posture and intention, enabling UAVs to perform more context-aware and purposeful actions beyond simple avoidance.

\vspace{-0.3cm}

\bibliography{Reference} 
\end{document}